# Probabilistic Exploration in Planning while Learning

Grigoris I. Karakoulas
Interactive Information Group
Institute for Information Technology
National Research Council Canada
Ottawa, Ontario, Canada K1A 0R6
grigoris@ai.iit.nrc.ca

## Abstract

Sequential decision tasks with incomplete information are characterized by the exploration problem; namely the trade-off between further exploration for learning more about the environment and immediate exploitation of the accrued information for decision-making. Within artificial intelligence, there has been an increasing interest in studying planning-while-learning algorithms for these decision tasks. In this paper we focus on the exploration problem in reinforcement learning and Q-learning in particular. The existing exploration strategies for Q-learning are of a heuristic nature and they exhibit limited scaleability in tasks with large (or infinite) state and action spaces. Efficient experimentation is needed for resolving uncertainties when possible plans are compared (i.e. exploration). The experimentation should be sufficient for selecting with statistical significance a locally optimal plan (i.e. exploitation). For this purpose, we develop a probabilistic hill-climbing algorithm that uses a statistical selection procedure to decide how much exploration is needed for selecting a plan which is, with arbitrarily high probability, arbitrarily close to a locally optimal one. Due to its generality the algorithm can be employed for the exploration strategy of robust Q-learning. An experiment on a relatively complex control task shows that the proposed exploration strategy performs better than a typical exploration strategy.

## 1 INTRODUCTION

Many decision-making tasks are inherently sequential since they are characterized by two features: incomplete knowledge and steps. These two features are interconnected. This is because most real-world decision problems occur within complex and uncertain environments in a continuous flow of events in time. Effective decision-making requires resolution of uncertainty as early as possible. The tendency to minimize losses resulting from wrong predictions of future events necessitates the division of the problem solution into steps. A decision at each step must make use of the information from the evolution of the events experienced thus far, but that evolution, in fact, depends on the type of decision made at each step.

Sequential decision tasks with incomplete information have long been studied in decision theory and control theory. Within artificial intelligence there has been increasing interest in studying these tasks, especially in the areas of planning (Dean & Wellman, 1991) and machine learning (Barto et al., 1989). In all these contexts, an agent that is given a goal to achieve in a partially known environment, plans its actions while learning enough about the environment in order to enable that goal. Such an agent should be able to represent and reason about change, uncertainty and the value or utility of its plans. Most importantly, though, it should be able to deal at any time with the trade-off between further exploration (also called identification, probing) and satisfactory exploitation (also called control, hypothesis selection, planning) of the accrued information. Several ideas for exploration strategies have been developed in the areas of statistical decision theory, optimal experiment design and adaptive control. There is currently an influx of these ideas into decision-theoretic planning (Russell & Wefald, 1991; Drapper et al., 1994; Pemberton & Korf, 1994), concept learning (Scott & Markovitch, 1993), speed-up learning (Gratch & DeJong, 1992; Greiner & Jurisica, 1992; Gratch et al., 1994), system identification (Cohn, 1994; Dean et al., 1995) and reinforcement learning (Thrun, 1992; Kaelbling, 1993).

In this paper we focus on exploration in reinforcement learning. The latter is a paradigm within machine learning appropriate to planning-while-learning tasks that has been shown to produce good solutions in domains such as games (Tesauro, 1992) and robotics (Mahadevan & Connell, 1993). The goal of reinforcement learning is to deter-mine a plan (i.e. a mapping from states of the environment into actions) that optimizes the expected value of a performance measure. An example of such a measure is the total long-term reward accrued from following a plan. The distribution of the performance measure of each plan depends on the dynamics of the environment which are assumed unknown. Furthermore, in tasks with large state and action spaces the search for an optimal plan within the space of possible plans is intractable. There is therefore need for efficient exploration that can be used to gather observations about the behavior of the environment. These observations should also be sufficient for selecting a plan which is probabilistically close to a locally optimal one.

The exploration strategies that have been developed for reinforcement learning are largely of a heuristic nature (Thrun, 1992). They also have limited scaleability in tasks with large (or infinite) state and action spaces. The main idea of this paper is as follows. Since in a planning-while-learning task an agent operates in a partially known environment, exploration should be guided by the effects of uncertainty on the performance estimates of plans. During exploration the agent can probe the environment to gather samples of state values. Our goal is to develop a probabilistic algorithm for deciding at each stage of the task how much sampling is needed for exploration in order that the plan selected at each stage be, with arbitrarily high probability, arbitrarily close to a locally optimal plan. The algorithm should be incremental so that its performance should monotonically improve with time as more computational resources are allocated to exploration.

## 1.1 Q-LEARNING FOR PLANNING WHILE LEARNING

Q-learning (Watkins, 1989) is the reinforcement learning algorithm that has been most studied both theoretically and practically. This is mainly due to its origination from the concepts and principles of dynamic programming (Bellman, 1957). Because of this relation with DP, Q-learning integrates planning and learning into a single algorithm in contrast to other reinforcement learning methods.

Let us define $X$ as the state space of the environment, $A$ as the action space and $P$ as the state transition model of the environment mapping elements of $X \times A$ into probability distributions over $X$. $r(x_t, a_t)$ is a reward function specifying the immediate reward that an agent receives by applying action $a_t$ at state $x_t$. A policy $\pi$ (i.e. a plan) is defined as $\pi: X \to A$. Given a policy $\pi$ from the set of possible policies $\Pi$, the value of an initial state $x_0$, $V_\pi(x_0)$, is the expected sum of rewards which are discounted by how far into the future they occur. Thus,

$$V_\pi(x_0) = E\left\{\sum_{t=0}^{\infty} \gamma^t r(x_t, a_t)\right\} \quad (1)$$

where $\gamma$, $0 < \gamma < 1$, is the discount factor and $a_t = \pi(x_t)$.

The Q-learning algorithm is based on the idea of maintaining for each state and action pair an estimate of $Q_\pi(x_t, a_t)$. The latter is an action-value function $Q_\pi: X \times A \to \Re$ that gives the expected discounted cumulative reward (reinforcement) for performing action $a_t$ in state $x_t$ and continuing with policy $\pi$ thereafter. According to the DP principle of optimality, the optimal value function $Q_{\pi^*}(x_t, a_t)$ can be written as:

$$Q_{\pi^*}(x_t, a_t) = E\{r(x_t, a_t) + \gamma V_{\pi^*}(x_{t+1})\} \quad (2)$$

where the expected value of the cumulative discounted reward from applying the optimal policy $\pi^*$ at state $x_{t+1}$ and thereafter, is given by

$$V_{\pi^*}(x_{t+1}) = max_{b \in A} Q_{\pi^*}(x_{t+1}, b) \quad (3)$$

If the state transition model is known, then the value iteration algorithm of DP can be applied so that at each iteration of the algorithm the Q-value of a state and action pair can be updated by

$$Q_{t+1}(x_t, a_t) = E\{r(x_t, a_t) + \gamma V_t(x_{t+1})\} \quad (4.1)$$

$$V_t(x_{t+1}) = max_{b \in A} Q_t(x_{t+1}, b) \quad (4.2)$$

where the expectation in (4.1) is over all possible next states. Successive iterations over the above two equations yield, in the limit, the optimal $Q_{\pi^*}$ function and hence the optimal policy $\pi^*$.

Unfortunately, the probability distributions of the state transition model are usually unknown. And even if these distributions were known, the task of identifying the optimal policy would be intractable (i.e. the "curse of dimensionality" (Bellman, 1957)).

In Q-learning the action model is assumed unknown. The agent only observes at each time step the value of the current state. This value could also be sampled from a random function. In that case, however, the agent does not need to know the stochastic characteristics of this sampling. The same applies to the value of the immediate reward, as this may also be determined probabilistically. The surface of the $Q_{\pi^*}$ function is learned by applying the rule

$$Q_{t+1}(x_t, a_t) = (1-\beta_t)Q_t(x_t, a_t) +$$
$$\beta_t [r(x_t, a_t) + \gamma V_t(x_{t+1})] \quad (5.1)$$

$$V_t(x_{t+1}) = max_a Q_t(x_{t+1}, a) \quad (5.2)$$

an infinite number of times to all possible state and action pairs. In (5.1) $\beta_t$, $0 < \beta_t < 1$, is the learning rate. The optimal policy $\pi^*$ is then obtained from the projection of the state space on the performance surface of $Q_{\pi^*}$. Learning of the $Q_{\pi^*}$ function can be intractable when the space $X \times A$ is too large for visiting all state and action pairs sufficiently enough. A simple exploration strategy like choosing an action according to a particular distribution (e.g. random walk, Boltzman distribution etc.) is inherently exponential especially in stochastic domains (Whitehead, 1991). There has been work on learning state transition models and/or utilizing knowledge generated from the Q-learning process in order to guide exploration (Sutton, 1990; Lin, 1991; Thrun, 1992). Although these exploration strategies have been shown to enhance the effectiveness of Q-learning, their efficiency can be questioned in complex tasks. This is because most of these strategies seek to perform exhaustive exploration. Furthermore, they do not provide any probabilistic guarantee for improvement of the policy being learned.

For this reason, we propose and develop a probabilistic exploration algorithm based on a selection procedure from sequential statistical analysis. Within each Q-learning iteration the algorithm uses this statistical procedure and the current estimates of the $Q_\pi$ functions to decide how much to explore within a possibly infinite set of policies. At the end of each iteration, the algorithm is probabilistically guaranteed to find a solution approximately close to a locally optimal one.

Section 2 presents the proposed exploration strategy. Section 3 shows how this strategy is incorporated into robust Q-learning, an algorithm specifically developed for adaptive planning in noisy and uncertain environments (Karakoulas, 1995a). Section 4 reports on an experiment for evaluating the performance of the proposed exploration strategy in robust Q-learning. Related and future work is discussed in Section 5. Conclusions are given in Section 6.

## 2  THE PROPOSED EXPLORATION STRATEGY

We assume that the agent has access to a stochastic discrete-time dynamic system that provides an approximation of the state of the environment at each time. The approximation need only be good enough for evaluating the relative performance of the policies. The system is in general of the form[1]

$$x_{t+1} = f(x_t, a_t, \theta, w_t) \quad (6)$$

where

$x_t$ is the state of the system at time $t$ and it may summarize past information that is relevant for future optimization;

$a_t$ is the action selected at time $t$ according to the policy function $a_t = \pi(x_t)$;

$\theta$ is the vector of parameters whose values are assumed unknown;

$w_t$ is a random parameter (also called disturbance or noise).

Actions can be discrete or real-valued. It should be noted that when the set of actions is infinite, the problem of searching for the best action in Q-learning becomes extremely difficult.

The agent does not know the probability distribution of the state of the system in (6). At any time the agent can apply an action and get a sample of possible next states as a result of the stochasticity of the system. It can then use this sample and the current estimate of the $Q_\pi$ function to statistically evaluate the likely effectiveness of the respective policy $\pi$. In standard Q-learning, whenever the agent applies an action to the environment it observes only one value of the next possible state. In our approach, on the other hand, the agent receives a sample of values of the next state through the partial model of (6). The latter, therefore, acts as an oracle for the agent.

Due to this deviation from standard Q-learning we introduce the notion of $q_\pi(x_t, a_t)$. The latter represents the expected discounted cumulative reward for performing action $a_t = \pi(x_t)$ in the state with particular value $x_t$ and continuing with policy $\pi$ thereafter. The $Q_\pi$ function can then be defined as the expected value of the distribution of q-values, i.e.

$$Q_\pi(x_t, a_t) = E\{q_\pi(x_t, a_t)\} \quad (7)$$

By this definition, the sum inside the brackets in (2) is equal to $q_{\pi^*}(x_t, a_t)$. The variance of the distribution of q-values is denoted by $\sigma_{q_\pi}^2$. Both the mean and the variance are assumed unknown. The agent can use samples of state values and the current estimates of q-values to compute estimates of the mean and variance. These estimates are denoted by $\overline{Q}_\pi$ and $S_{q_\pi}^2$ respectively.

---

[1] When the state of the environment is discrete, as it is usually the case with AI planning problems, the stochastic process of the state of the environment can be approximated by the state transition model.

During exploration the agent seeks a policy $\pi' \in \Pi$ such that

$$\pi' = \text{argmax}_{\pi \in \Pi} Q_\pi(x_t, a_t) \quad (8)$$

The policy function $\pi(x_t)$ is assumed to be defined in terms of a set of parameters c. In the case of real-valued actions the function can be for example of the linear form $\pi(x_t) = c \cdot x_t$. The values of the parameters are unknown. An instance of parameter values identifies a particular policy $\pi$. As the set of possible policies $\Pi$ is infinite, the common solution for pruning the space of alternatives is the search technique of hill-climbing. In particular, the technique of steepest-ascent hill-climbing can climb the gradient of Q-values in (7) by selecting the policy having the highest Q-value with respect to the current policy. An apparent limitation of this approach is that it requires the probability density function of $x_t$ to be expressed in an analytic form. Even when such information is available its use in this type of search makes the problem computationally intractable. To overcome this serious limitation we follow an empirical approach to the exploration problem.

As already mentioned samples of state values can be used to derive estimates of the mean and variance of the q-value distribution of a policy $\pi$. A probabilistic algorithm can be built which, given the samples, can as efficiently as possible hill-climb from an initial policy $\pi$ to one that is, with high probability, a local optimum. The search is performed using a set of transformations of an initial policy $\pi$, $T = \{T_i(\pi)\}$. Each $T_i$ maps a policy $\pi$ into another policy $\pi'$. Such mapping can be performed for example through a perturbation of parameter values of the policy function at the point identified by $\pi$.

Efficiency of the search refers to the bounded number of samples that are sufficient for the probabilistic algorithm to output a solution with statistical significance. Such efficiency can be achieved by incorporating a sequential statistical procedure (Govindarajulu, 1987) into the hill-climbing algorithm. Our interest in such procedures is due to their ability to reach an inference earlier than a fixed-sample size procedure. In the latter, the size of the sample is fixed prior to any statistical experiment. The distinct characteristic of a sequential procedure is that the number of observations required to terminate a statistical experiment is a random variable as it depends on the outcome of the observations from the experiment. Inference in sequential statistical procedures is performed via a statistical test called a *stopping rule*. This rule determines the sufficient number of observations that need to be made in order that the null hypothesis of the statistical test is rejected with a specific degree of error. The number of observations that have been made when the stopping rule is satisfied, is called the *stopping time*. A sequential statistical procedure can, therefore, meet our requirement for an incremental exploration algorithm.

Let us define the *local policy improvement operator* $I : \pi \rightarrow \pi$ by

$$I(\pi) = \begin{cases} T_i(\pi) & \text{if } Q_i(x_t, a_t) - Q_\pi(x_t, \pi(x_t)) > 0 \\ \pi & \text{otherwise} \end{cases} \quad (9)$$

where $T_i(\pi)$ is policy transformation $i$ from a countably infinite set $T = \{T_i(\pi)\}$ in a neighborhood of the current policy $\pi$. $Q_i$ denotes the Q function of policy transformation $T_i(\pi)$. Also $a_t = T_i(\pi(x_t))$.

Because of the stochasticity of the environment, the inequality between the Q-values in (9) can only be satisfied with a particular level of statistical confidence. Since no probability density functions are assumed, this inequality should be empirically assessed from the random samples of states. For this purpose we next introduce the sequential selection procedure by Dudewicz and Dalal (Dudewicz & Dalal, 1975).

Consider the problem of selecting from $k$ populations $(k \geq 2)$ — each being distributed as $N(\mu_i, \sigma_i^2)$ with $\mu_i$ and $\sigma_i^2$ unknown — the population that has the largest mean. The selection of the largest mean is done with probability at least $p^*$ whenever the difference between the top two means is at least equal to some value $\varepsilon$. That is

$P\{\text{correct selection}\} \geq p^*$

$$\text{if } (\mu_{[k]} - \mu_{[k-1]}) \geq \varepsilon \quad (10)$$

where $\mu_{[1]} \leq \ldots \leq \mu_{[k]}$ denotes the ordered sequence of means and $p^* = 1 - \delta$ with $\delta$ being the error probability.

In our case a population $i$ corresponds to the population of mean Q-values, $\{\overline{Q}_i\}$, of policy transformation $T_i(\pi)$, $i = 1, \ldots, k$. Each of the mean Q-values is estimated by

$$\overline{Q}_i = \frac{1}{N} \sum_{j=1}^{N} q_{ij} \quad (11)$$

where $N$ is the size of a sample of state values and hence of a sample of q-values. Each $\overline{Q}_i$ is an unbiased estimator of $Q_i$. From mathematical statistics we have that for any random variable y

$$\hat{\mu} = \bar{y} \sim N(\mu, \sigma^2/N) \quad (12)$$

The sequential statistical selection problem of (10) suits our purpose of selecting from a set of $k$ populations, $\{\overline{Q}_i\}$ $i = 1, ..., k$, the one with the highest Q-value, with probability of correct selection at least $p^*$. Since (10) cannot be satisfied by a sequential procedure that involves only a single stage of sampling, Dudewicz and Dalal have constructed a two-stage procedure for determining the minimum size of the sample for each population. The procedure is based on a multivariate t-distribution for defining the probability $p^*$ of correct selection in (10).

This sequential selection procedure can form the basis of a probabilistic hill-climbing algorithm. When a policy is selected at the end of an iteration of the algorithm, the procedure is again applied in the next iteration for a set of local transformations of the newly selected policy. The search terminates at the iteration where the selection procedure selects the same policy as at the last iteration. That policy is a probably locally optimal one. The error of each stage $\delta_i$ can be set such that the total error over all stages $\delta$ is less than some pre-specified constant.

The probabilistic hill-climbing (PHC) algorithm is presented in Figure 1. When the algorithm is invoked for exploration during a Q-learning iteration it is initialized with the policy of the previous iteration. For ease of notation we denote the mean of q-values, $\overline{Q}_i$ defined in (11), with $\eta_i$. $\omega$ counts the iterations of the PHC algorithm. The value of $\varepsilon_\omega$ is dynamically determined for each set of transformations $T$ according to the values of the policy-improvement operators of the set. The symbol $\lceil . \rceil$ denotes the smallest integer greater than or equal to the quantity enclosed. The values of $h$ are given by tables in Dudewicz and Dalal (1975). Specific values for the $\tau_{ij}$ are also suggested in that paper. At each iteration of the algorithm the selection procedure starts with $n_0$, $(n_0 \geq 2)$, samples $\eta_{i1}, ..., \eta_{in_0}$ from each population $\{\overline{Q}_i\}$. Additional samples are taken according to the stopping time $n_i$ of each population. The policy transformation $T_{[k]}(\pi)$ with the maximum $\eta_{[k]}$ value is then selected. The probability of this selection being the correct one is:

$$P\{Q_{[k]} - Q_{[k-1]} \geq \varepsilon\} \geq 1 - \delta \quad (13)$$

If several top $Q_i$ values are less than $\varepsilon$-close, the above procedure may not select the policy with the highest Q-value. The amount of error from the selection depends on $\varepsilon$. In this case the selection of a policy that is not the best one, is not, however, a drawback since the goal of PHC algorithm is to explore. In fact, the parameters $\varepsilon$ and $\delta$ of

---

Algorithm PHC $(\pi, \Psi, N, \delta)$

$\omega \leftarrow 0$;

**While** there is a set of policy transformations $\{T_i(\pi)\}$ **do**

  **For** each policy transformation $T_i(\pi)$ in the set **do**

    Take $n_0$ samples $\eta_{i1}, ..., \eta_{in_0}$ from population $\{\overline{Q}_i\}$;

    Calculate the variance of the $n_0$ samples

$$S_i^2 = (1/(n_0-1)) \cdot \sum_{j=1}^{n_0} (\eta_{ij} - \overline{\eta}_i);$$

    Calculate stopping time $n_i = max\left\{n_0+1, \left\lceil \left(\frac{S_i h}{\varepsilon_\omega}\right)^2 \right\rceil \right\}$;

    Take $n_i - n_0$ additional samples of $\eta_{ij}$;

    Calculate $\overline{\eta}_i = \sum_{j=1}^{n_i} \tau_{ij} \cdot \eta_{ij}$.

  Choose the policy transformation $T_{[k]}(\pi)$ with $k = argmax_T(\overline{\eta}_i)$;

  $\pi \leftarrow T_{[k]}(\pi)$;

  $\omega \leftarrow \omega + 1$;

Return as output policy $\pi$.

Figure 1: The probabilistic hill-climbing (PHC) algorithm.

---

the algorithm can be used to control exploration according to the degree of learning.

## 3 EXPLORATION IN ROBUST Q-LEARNING

In a planning task an agent should be able to reason about uncertainty in its model of the environment as well as about the effects of that uncertainty on finding a satisfactory plan. The concept of robustness refers to three issues that capture the effects of uncertainty on the performance of a plan. These issues are: (i) the stability of the environment's behavior under the plan, (ii) the expected total reward of the plan and (iii) the variability in the total reward as an indicator of sensitivity to uncertainty. Reasoning about the three issues can be done by evaluating the agent's attitude to the risk that is involved when following the course of action of a particular plan within the partially known environment.

In general, risk can be considered as one's willingness to bet against the odds of a chance prospect (e.g. a lottery).

We adopt the concept of risk aversion as stated by Diamond and Stiglitz (1974) in order to construct a measure of robustness that reflects the agent's attitude to the risk associated with a plan. Assuming a utility function with a constant absolute risk-aversion parameter we derive the following utility measure

$$\tilde{U}_\pi = \varphi \overline{U}_\pi - (1 - \varphi) \sigma_\pi^2 \quad (14)$$

where $U_\pi$ is the total discounted reward from policy $\pi$, $\overline{U}_\pi$ is its mean, $\sigma_\pi^2$ is its variance and $\varphi$, $0 < \varphi < 1$, is the risk-aversion parameter. According to (14), in a situation of increasing risk where the mean value of $U_\pi$ is preserved but its variance is increased (a mean-preserving increase in risk), a risk-averse agent would feel worse-off by a degree equal to $(1 - \varphi)/\varphi$.

Using (14) we can build a robust Q-learning algorithm (Karakoulas, 1995a) in which the reward at each iteration is defined as

$$\tilde{R}_t(x_t, a_t) = \varphi \overline{R}_t - (1 - \varphi) S_{r_t}^2 \quad (15)$$

where $\overline{R}_t$ and $S_{r_t}^2$ are the mean and variance of the immediate reward from applying action $a_t$ to the environment. They are estimated from a sample of states of the environment. The counterparts of equations (5.1) and (5.2) for updating the Q-values are

$$q_{t+1}(x_t, a_t) = q_t(x_t, a_t) + \beta_t P(x_t) \cdot$$
$$\left[ \tilde{R}(x_t, a_t) + \gamma \tilde{V}_t(x_{t+1}) - q_t(x_t, a_t) \right] \quad (16.1)$$

$$\tilde{V}_t(x_{t+1}) = \max_{\pi' \in T(\pi)} \quad (16.2)$$
$$\left[ \tilde{Q}_t(x_{t+1}, \pi'(x_{t+1})) = \varphi \overline{Q}_t - (1 - \varphi) S_{q_t}^2 \right]$$

In (16.1) the probabilities $P(x_t)$ are estimated from the sample of states at time $t$ using Bayes' rule. Equations (16.1) and (16.2) define the updating rule of robust Q-learning.

To see whether the PHC algorithm can be employed in robust Q-learning, we write the formula of the $Q$ function in (16.2) in terms of a random variable with expected value $\mu$ and variance $\sigma^2$

$$\varphi \mu - (1 - \varphi) \sigma^2 \quad (17)$$

It can be shown that the random variable $\eta$,

$$\eta = \varphi \hat{\mu} - (1 - \varphi) \hat{\sigma}^2 \quad (18)$$

---

1. Initialize;
2. **For all t do:**
    (i) Create a sample $S_t$ of current instances of states;
    (ii) Search probabilistically via PHC for the locally optimal policy $\pi$;
    (iii) Apply policy $\pi$ to the sample $S_t$; obtain new sample $S_{t+1}$;
    (iv) Estimate the reward $\tilde{R}$ from the sample by (15);
    (v) Update the $\tilde{Q}$ value of sample $S_t$ and policy $\pi$:
        update the q-value of each instance in the sample by (16.1)-(16.2);
        match instance to clusters of policy $\pi$;
        merge instance if matching conditions satisfied;
    (vi) merge existing clusters of $\pi$ that satisfy matching conditions.

---

Figure 2: The steps of the robust Q-learning algorithm.

where $\hat{\mu}$ and $\hat{\sigma}^2$ are sample estimates of the mean and variance, is normally distributed with mean and variance that depend on $\mu$ and $\sigma^2$. Hence, they are assumed unknown.

We can therefore apply the PHC algorithm as given in Figure 1 by using (18) for the definition of the random variable $\eta_i$ of a policy transformation $T_i(\pi)$. Since $\eta_i$ is an unbiased estimator of $Q_i$, within each iteration of PHC the sequential statistical procedure finds the minimum number of observations for each $\eta_i$ and selects the policy with the highest $Q$-value at a particular level of statistical significance.

For completeness of exposition we present the basic steps of the robust Q-learning algorithm in Figure 2 (for more details see (Karakoulas, 1993; 1995)). The steps (v) and (vi) of the algorithm refer to the function approximator that is used for generalization of the q-values over real-valued state and action spaces. For each policy clusters are formed to approximate the Q function of that policy. Given a sample of states, the q-values of the sample under a particular policy are estimated by matching each state of the sample with the states already stored in the clusters of the policy.

## 4 EXPERIMENTAL RESULTS

The purpose of the experiments reported in this section is to demonstrate the effectiveness of the PHC algorithm that has been implemented as the exploration strategy of robust Q-learning. Thus, we have run two experiments, in one applying this exploration strategy and in the other applying the simple and most used semi-uniform exploration strategy. The latter strategy chooses, at each time, the policy with the highest $Q$-value with a predefined probability $\xi$, and a random policy with probability $1-\xi$.

The experiments are performed on an adaptive control task in which the environment is approximated by the following partially known model[2]

$$x_{1,t+1} = \kappa \cdot x_{1,t} + (1-\kappa) x_{3,t} \quad (19.1)$$

$$x_{2,t+1} = (1-\vartheta) x_{2,t} + \vartheta \cdot \zeta \cdot a_t \quad (19.2)$$

$$x_{3,t+1} = x_{1,t+1} + \upsilon \cdot x_{2,t+1} \quad (19.3)$$

The state of the model is a vector of three variables. There is only one action variable $a_t$ which the agent can use in order to control the state of the model. There is uncertainty in the model since the exact values of the parameters of the model $\kappa$, $\vartheta$, $\zeta$ and $\upsilon$ are assumed unknown. The parameters get random values from uniform distributions in [0.6,0.9], [0.1,0.4], [0.4,0.6] and [1.5,2.5] respectively. At any time $t$ values of the state of the environment are computed by applying Monte Carlo simulations. In these simulations the parameter values are randomized according to the aforementioned distributions.

The control task of the agent is: given a shock upon the environment through the state variable $x_{1,t}$, find the optimal policy that drives the environment back to its initial state. In adaptive control theory, the control task in linear systems with uncertain parameters — such as the one of this experiment — is usually considered as a non-linear stochastic control task. This is because a closed-form solution of the optimal policy does not generally exist. For this reason, good suboptimal policies are sought in practice. In addition, in such task the trade-off between exploration and exploitation is crucial for finding a good solution (Kumar, 1985).

Let us assume that the initial values of the state variables are zero. The goal of the agent must be reflected in its utility function that penalizes whenever either $x_{2,t}$ or $x_{3,t}$ deviates from its initial value. The utility (or cost) function is assumed to be of the form

$$U = \sum_{t=0}^{\infty} \gamma^t \left[ \tau_1 x_{2,t}^2 + \tau_2 x_{3,t}^2 \right] \quad (20)$$

where the coefficients $\tau_1$ and $\tau_2$ have negative values for transforming the original minimization problem into a maximization one. The action variable $a_t$ in this task is defined by the following policy

$$a_t = \pi(x_{1,t}) \quad (21)$$

We assume that the policy is a linear function

$$a_t = c \cdot x_{1,t} \quad (22)$$

Such linear policy functions are common in optimal control problems because they are robust under uncertainty and they are easy to implement.

In the two experiments, the coefficients in (20) had values $\tau_1 = -5$ and $\tau_2 = -5$. The discount rate was set to 0.988. The risk-aversion parameter $\varphi$ in (15) and (16.2) was set to 0.5 giving an equal weight to both the mean and variance. In the first experiment, the parameters of the PHC algorithm were set to $\pi = 0$ (i.e. no policy initially assumed), $N$ was set to 50 and $\delta$ was set to 0.04. Because of the linear policy function in (22) a policy transformation $T_i(\pi)$ generates a new policy from a policy $\pi$ by moving the gradient of (22) by a small step.

In the second experiment the parameter of the semi-uniform exploration strategy $\xi$ was set to 0.1. Thus, the best policy was chosen with probability 0.9 and any other policy in the set of possible policies was chosen with probability 0.1. To enable this randomization in policy selection we constructed a finite subset of policies from the original infinite set. The policies in the subset were defined by (22) with coefficient values in the discretized range -2.4,-2.39,...,0.29,0.3.

The results of the two experiments are presented in Figures 3 and 4. The curves from the learning algorithm with the semi-uniform distribution are depicted with dashed lines. The experiments were run for 30 time-periods. Both learning algorithms converged to the optimal policy $a_t = (-0.69) \cdot x_{1,t}$. This is the same policy that was found by Kemball-Cook (1993). He used a control theory approach for solving this problem. Figure 3 shows the convergence of the two algorithms to the policy rule as a percentage of the learning run. Figure 4 shows the convergence of the algorithms in terms of the cumulative reward obtained from following the learned policy averaged over the number of times this policy was active. In both figures the learning algorithm with the PHC exploration converges faster than the algorithm with the semi-uniform exploration by a factor of three. This seems to be in agreement with empirical studies in machine learning that have demonstrated drastic reduction in the number of training experiences when the exploration strategies for selecting training experience use information about the current state of the search of the theory space (see (Scott and Markovitch, 1993) for more on this).

---

[2] Both the model and the control task have been of particular interest in the field of economic dynamics and control (Kemball-Cook, 1993; Karakoulas, 1993). We present them here by abstracting them from any domain-specific details.

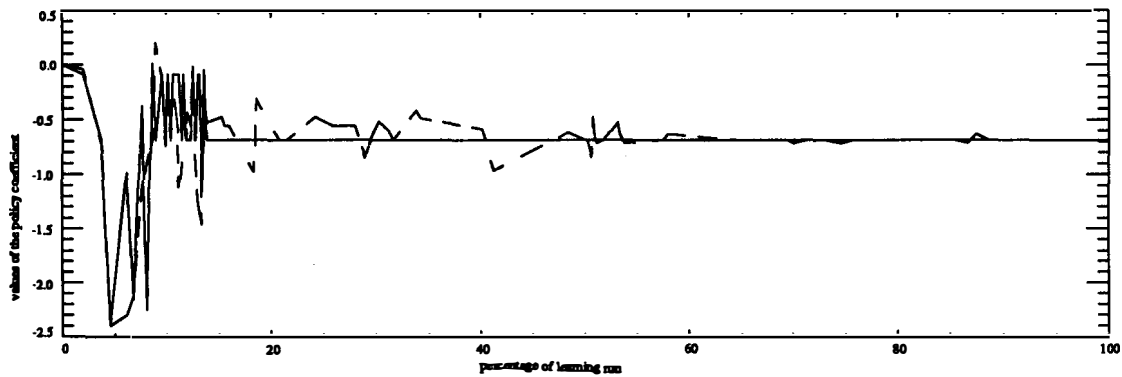

Figure 1: Convergence to the optimal policy

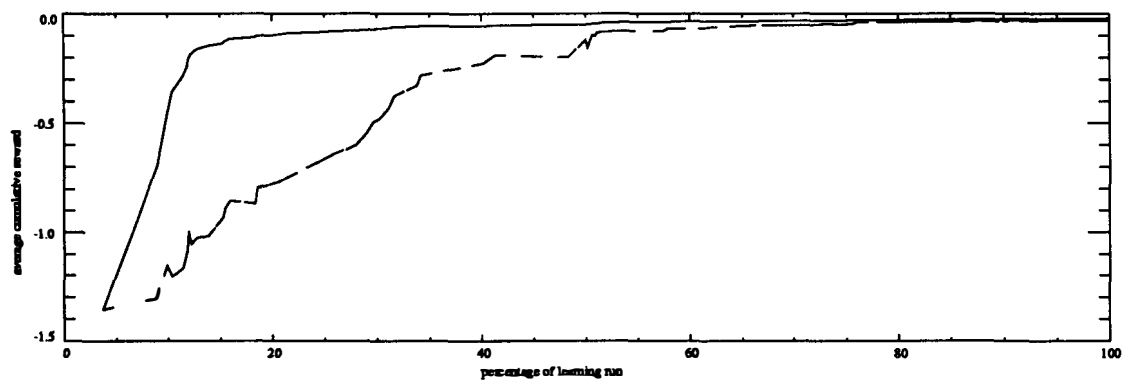

Figure 2: Convergence of the average cumulative reward from the optimal policy

## 5 DISCUSSION

The exploration strategies that have been developed for speed-up learning (Gratch & DeJong, 1992; Greiner & Jurisica, 1992; Gratch et al., 1994) are also based on sequential statistical analysis. The selection procedures involved are of only one stage and they are not therefore appropriate for our selection problem which requires a two- or a multi-stage procedure.

Kaelbling (1993) has developed a statistical algorithm for exploration in reinforcement learning. The algorithm works by keeping statistics on the number of times a given action has been executed and the proportion of times that it has succeeded in terms of yielding a fixed reward. Based on these statistics the algorithm constructs confidence intervals for the expected reward for each of the feasible actions; and uses the upper bound of the intervals for choosing actions and for updating the estimates of value functions. The confidence intervals are estimated from the standard normal distribution or from non-parametric statistical techniques. In these interval estimation procedures the size of the sample is predefined whereas in PHC it is determined on-line according to the observations gathered through exploration. This enables the PHC algorithm to efficiently perform experimentation and to exploit this experimentation for finding with statistical significance a locally optimal policy. In addition, the PHC algorithm can handle real-valued actions. The relative performance of the two algorithms needs to be empirically evaluated.

In decision-theoretic planning Pemberton and Korf (1994) have proposed separate heuristic functions for exploration and decision-making in incremental real-time search algorithms. Draper et al. (1994) have developed a probabilistic planning algorithm that performs both information-producing actions and contingent planning actions. Our exploration strategy could be applied to these planning tasks as part of a Q-learning algorithm. Of course, the search space and the transformation operators of PHC must be appropriately defined in terms of the actions of each task. We plan to examine the performance of the PHC algorithm within a Q-learning algorithm that has recently been developed for the task of cost-effective classification (Karakoulas, 1995b). This is a planning-while-learning task in which the exploratory actions (e.g. diagnostic tests) have a cost associated with them.

The PHC algorithm is related in principle to the fully-polynomial randomized approximation schemes that have been developed for approximating solutions of enumeration and reliability problems (Karp & Luby, 1983; Jerrum & Sinclair, 1988). These problems are in general intractable. The algorithms run in time polynomial in the size of the search space and output an estimate of the solution which is, with high probability, $\varepsilon$-close to the solution. Jerrum and Sinclair (1988) envisage the application of their algorithm to the process of simulated annealing. This process has been used in combination with the Boltzman distribution for controlling exploration in reinforcement learning.

In work related to our robust Q-learning method, Heger (1994) has proposed a Q-learning algorithm based on the minimax criterion. The latter defines the most risk-averse control strategy. In contrast, in our approach the risk-aversion parameter is used to trade-off the most risk-averse criterion represented by the variance of q-values with the most risk-neutral criterion represented by the expectation of q-values. Hence, different types of risk-averse strategies can be realized by appropriately setting the value of the risk-aversion parameter. It is worth pointing out that our robustness criterion considers only the variance of q-values due to sampling error. We plan to extend this criterion by including the bias factor due to the estimation error.

## 6 CONCLUSION

In this paper we have examined the problem of exploration that occurs when applying Q-learning for planning-while-learning tasks in uncertain environments. We proposed a strategy that uses information about the effects of uncertainty on the evaluation of alternative policies in order to guide exploration in Q-learning. A probabilistic hill-climbing (PHC) algorithm was developed for implementing the strategy. The algorithm iterates over a two-stage sequential statistical procedure that finds the minimum number of observations required for selecting a locally optimal policy with a particular level of statistical confidence. The sequential procedure makes the algorithm incremental. Furthermore, the assumptions of the procedure do not impose any restrictions on its applicability in Q-learning. For this reason, we were able to incorporate the procedure in robust Q-learning which is a Q-learning algorithm based on risk-averse Q functions. The effectiveness of the exploration strategy was tested by applying robust Q-learning on a realistic adaptive control task. Two experiments were performed for comparing the performance of the learning algorithm using PHC and using the typical semi-uniform distribution. The learning algorithm with PHC converged faster by a factor of three. Future work will examine the applicability of the exploration strategy in the planning-while-learning task of cost-effective classification.

### Acknowledgements

Thanks to Martin Brooks, Innes Ferguson and the anonymous referees for their very helpful comments on earlier versions of this paper.